\documentclass[letterpaper,10pt]{article} 
\usepackage{osameet3} %% use version 3 for proper copyright statement

\usepackage{amsmath}
\usepackage{url,hyperref,lineno,microtype,subcaption}
\usepackage[onehalfspacing]{setspace}
\usepackage{amsthm}
\usepackage{amssymb}
\usepackage{array}
\usepackage{tabularx}
\usepackage{booktabs}
\usepackage{pifont}
\usepackage{footnote}
\newcommand{\xmark}{\ding{55}}%

\newcommand{\secref}[1]{Section~\ref{#1}}

\newcommand{\tabref}[1]{Table~\ref{#1}}

\newcommand{\defref}[1]{Def. \ref{#1}}

\newcommand{\g}[1]{g_{#1}}

\newcommand{\gnum}{M}
\newcommand{\gmap}{\mathcal{G}}
\newcommand{\w}[1]{\omega_{#1}}

\newcommand{\wg}[1]{\omega_{g,#1}}
\newcommand{\we}{\omega_e}
\newcommand{\p}[1]{p_{#1}}
\newcommand{\pset}{{\bf p}}
\newcommand{\pnum}{N}
\newcommand{\sethree}{SE(3)}

\newtheorem{definition}{Definition}

\begin{document}

\title{Robotic Grasping from Classical to Modern: A Survey \footnote{This draft will be continuously updated. Therefore, if you find any problems with this draft, please do not hesitate to contact the first author for updates, including but not limited to: 1) other interesting works or ideas not included in this draft; 2) problems with the statements in this draft; 3) further discussion about the included works; 4) other useful suggestions.}}

\author{Hanbo Zhang\,$^{1}$, Jian Tang\,$^{1}$, Shiguang Sun\,$^{1}$ and Xuguang Lan$^{1}$}
\address{Xi'an Jiaotong University}
\email{zhanghanbo163@stu.xjtu.edu.cn}
%%Uncomment the following line to override copyright year from the default current year.
\copyrightyear{2022}

\begin{abstract}
Robotic Grasping has always been an active topic in robotics since grasping is one of the fundamental but most challenging skills of robots.
It demands the coordination of robotic perception, planning, and control for robustness and intelligence. 
However, current solutions are still far behind humans, especially when confronting unstructured scenarios. 
In this paper, we survey the advances of robotic grasping, starting from the classical formulations and solutions to the modern ones. 
By reviewing the history of robotic grasping, we want to provide a complete view of this community, and perhaps inspire the combination and fusion of different ideas, which we think would be helpful to touch and explore the essence of robotic grasping problems. 
In detail, we firstly give an overview of the analytic methods for robotic grasping. 
After that, we provide a discussion on the recent state-of-the-art data-driven grasping approaches rising in recent years. 
With the development of computer vision, semantic grasping is being widely investigated and can be the basis of intelligent manipulation and skill learning for autonomous robotic systems in the future. 
Therefore, in our survey, we also briefly review the recent progress in this topic. 
Finally, we discuss the open problems and the future research directions that may be important for the human-level robustness, autonomy, and intelligence of robots.

\end{abstract}

\section{Introduction}

% Motivation
With the development of robotics, robots are gradually entering our homes. Before being capable of sophisticated daily tasks, robots must be firstly seasoned in basic skills, among which grasping should be the most important one. To perform robust grasping, perception, planning, and control are simultaneously required. Therefore, robotic grasping is a fundamental but most challenging area in robotics.
Though actively investigated for several decades, robotic grasping is far behind being solved, especially when the robot is confronting complex and unstructured environments, or demanded to perform tasks with high-level semantics, which, however, is what we always hope an intelligent robot helper to be capable of.
Fortunately, with the development of deep learning~\cite{lecun2015deep}, there is remarkable progress in the last decade for learning representations of high-level semantics, such as object recognition and relationship understanding~\cite{liu2020deep}, natural language understanding~\cite{otter2020survey}, and robotic skill learning~\cite{kroemer2021review}.
By taking advantage of the recent progress in learning, it is promising to build an intelligent robot that could percept and understand the world as a human can do, interact with humans using natural languages, and finish grasping tasks autonomously and robustly with the abstracted semantics, serving as the basis of achieving more complicated tasks.

% Purpose
Therefore, in this paper, we will review the recent advances achieved in robotic grasping in the past several years, starting from the classical formulations to modern society.
By this survey, we want to answer the following questions:
\begin{itemize}
	\item Mathematically, what is grasping?
	\item How can we solve the problem of grasping?
	\item What are the advantages and disadvantages of the existing methods?
	\item What could the future trends and directions in this field?
\end{itemize}
Obviously, it is impossible to include all works related to robotic grasping in one paper. Therefore, according to the above questions, we hope to select a representative subset of this field, and provide the readers a comprehensive and well-organized overview from formulation to solutions.

In brief, to answer the above questions and find feasible solutions, researchers have explored for decades, and it results in an extensive set of excellent works. In detail, in the early stage, most works focused on the analytic form of grasp synthesis based on mechanics, e.g., force-closure and form-closure \cite{bicchi1995closure} grasp synthesis. 
However, such methods always rely on the simplification of the physical models and the assumption of a fully-observable environment, which could be hardly achieved in real-world scenarios. 
With the rapid development of learning approaches, data-driven approaches gradually dominated the community since it was simple, efficient, and could get rid of the strong assumptions made by the analytic approaches \cite{bohg2013data}. 
Nevertheless, data-driven methods are always data-intensive, meaning that they usually require much more data for training the grasping policy, which is always labor-intensive.
To solve the problem of data, self-supervised learning and unsupervised learning are extensively explored in recent years \cite{jing2020self, bengio2012unsupervised}, including some excellent works in robotic grasping \cite{mar2015self, pinto2016supersizing, zeng2018learning, berscheid2019improving, berscheid2020self}.
It is also possible to train grasping policies in physical simulators and then transfer to the real world \cite{yan2017sim, james2019sim, iqbal2020toward, yan2019data, zhao2021regnet, zhang2022regrad}.
By enough data, the performance of data-driven approaches substantially outperforms the classical methods.
Based on the current progress, there are several interesting questions:
\begin{itemize}
	\item Grasping is essentially a physical action, and hence, the classical analytic methods are well motivated. Therefore, {\it could we take the best of both analytic and data-driven to develop robust and scalable grasping methods?}
	\item Rapid development in computer vision reveals that large-scale learning could potentially abstract the internal structure of complex data. {\it Could we take advantage of the vision techniques for developing robust grasping skills in semantic scenarios?}
	\item The real world is full of uncertainty. Failing to handle uncertainty will severely affect the reliability and robustness, limiting the practicality. {\it Could we model the uncertainty from the learned models when planning grasps for robustness?}
\end{itemize}
We will discuss all the above questions in detail in this survey.

\begin{table}[t]
\caption{Robotic Grasping Surveys in the Last Few Decades}
\label{table:surveys}
\begin{center}
\begin{tabularx}{1\columnwidth}{lp{4.3cm}<{\centering}p{8.2cm}}
\toprule
 \bf Author \& \bf Year & \bf Type & \bf Summary \\
\midrule
\cite{shimoga1996robot} & Analytic Grasping & Analyze grasping approaches in the aspects of dexterity, equilibrium, stability, and dynamic behaviors based on mechanics.\\
\specialrule{0em}{2pt}{2pt}
\cite{bicchi2000robotic} & Analytic Grasping & An educational survey on analytic grasping, introducing all key parts for analytically synthesizing mechanically robust grasps.\\
\specialrule{0em}{2pt}{2pt}
\cite{sahbani2012overview} & Analytic and Data-driven Grasping & A short survey including both analytic and data-driven approaches.\\
\specialrule{0em}{2pt}{2pt}
\cite{bohg2013data} & Data-driven Grasping & Study the data-driven methods, mainly focusing on grasping using traditional machine learning techniques.\\
\specialrule{0em}{2pt}{2pt}
\cite{elango2015review} & Soft-hand Manipulation & A survey on soft-hand design and manipulation, including both hardware and software parts. \\
\specialrule{0em}{2pt}{2pt}
\cite{roa2015grasp} & Grasp Evaluation & Focus on grasp quality evaluation methods, and classify them into {\it contact-point-based} and {\it hand-configuration-based} metrics.\\
\specialrule{0em}{2pt}{2pt}
\cite{huang2016recent} & Datasets & Summarize the datasets before 2016 related to robotic manipulation, including grasping.\\
\specialrule{0em}{2pt}{2pt}
\cite{kleeberger2020survey} & Deep-learning Grasping & Categorize grasping methods into {\it model-based} and {\it model-free}, and focus on the recent deep-learning methods. \\
\specialrule{0em}{2pt}{2pt}
\cite{cong2021comprehensive} & Visual Manipulation & A short survey on vision-based manipulation, especially grasping approaches.\\
\specialrule{0em}{2pt}{2pt}
\cite{kroemer2021review} & Data-driven Manipulation & A comprehensive survey on learning-based robotic manipulation, mainly focusing on the (PO)MDP formulation. \\
\midrule
Ours (2022) & Analytic, Data-driven, and Object-centric Grasping & A survey to review the progress in the last few decades, including the traditional analytic approaches, recent data-driven approaches, and arising object-centric approaches. \\

\bottomrule
\end{tabularx}
\end{center}
\vspace{-15pt}
\end{table}

% Other surveys
There are also some other reviews on the topic of robotic grasp synthesis.
For example, \cite{bohg2013data} and \cite{kleeberger2020survey} have reviewed the recent data-driven grasping approaches with different taxonomies. 
To be more specific, \cite{bohg2013data} mainly discussed the methods based on classical machine learning techniques by categorizing them into methods for {\it known objects}, {\it familiar objects}, and {\it unknown objects}. 
By contrast, \cite{kleeberger2020survey} focused more on the recent deep learning methods.
More generally, \cite{kroemer2021review} and \cite{cong2021comprehensive} reviewed the learning-based robotic manipulation, in which the grasping skill is certainly included.
On the other hand, \cite{sahbani2012overview}, \cite{bicchi2000robotic}, and \cite{shimoga1996robot} surveyed the analytic grasp synthesis approaches in different years.
Besides, as for the grasp quality evaluation methods, one can refer to \cite{roa2015grasp} for a comprehensive study.
Different from them all, our paper mainly focuses on the topic of semantic grasping, and only includes the necessary and representative backgrounds of the traditional grasp synthesis.
In \tabref{table:surveys}, we also provide a comparison among all these surveys for the readers to choose the interested ones.

% Organization
Our survey is organized as follows. 
In \secref{sec:formulation}, we give a short overview of the mainstream formulations of robotic grasp synthesis. Generally speaking, a specific formulation usually accords with a specific kind of grasping approach.
In \secref{sec:analytic-grasp}, we will review a series of representative and impactful works related to analytic grasp synthesis along with the mechanics-based grasp quality evaluation methods, which inspired later works and formed a basis for the robotic grasping community. 
In \secref{sec:data-driven-grasp}, we will discuss the data-driven grasping approaches, aiming at synthesizing grasps from experiences, which are usually represented by a dataset. 
In \secref{sec:semantic-grasp}, we are going to survey the object-centric grasping approaches, which are usually targeted at a certain object specified by humans possibly with different interfaces such as a class name or a natural language command. Also, object-centric grasp planning in dense clutter involves the understanding of object relationships, which is also discussed in this section.
Finally, in \secref{sec:future}, we will discuss the open problems of robotic grasping that are important but still remain unsolved, and the future trends in these areas.

%\section{Taxonomy of Robotic Grasping}
%\label{sec:taxonomy}

\section{Problem Formulation}
\label{sec:formulation}

Grasp synthesis in robotics means finding the proper configuration of the robot's actuator related to the state of the target for stable grasping.
It could be formalized in different ways.
Concretely, the classical formulation focuses more on the mechanical properties, while modern approaches show more interest in the visual properties.
In this section, we will review different formulations of grasp synthesis.

\subsection{Overview}
Basically, given an object represented by 2D- or 3D-format, it would be a challenging problem to find an optimal, or at least stable, grasp from infinite candidates based on the geometric or physical analysis.
Therefore, to develop robust grasping approaches, several challenges will instruct the discussion in this section:
\begin{itemize}
	\item How can we properly represent a grasp?
	\item How can we evaluate the quality of a given grasp?
	\item How can we efficiently sample high-quality grasp candidates from an infinite set?
\end{itemize}
In this section, we will discuss the first two questions, and leave the answer to the third question as to the main body of this paper. Noticeably, there is another important issue about how to plan to execute the optimized grasp given the kinematics of the actuator without collision with the environment, which is also crucial for a successful grasping trial. However, it mostly relates to motion planning algorithms \cite{lavalle2006planning}, and goes out of the scope of this paper.

\subsection{Grasp Representation}

% Mathematical Formulation and Assumption
Before reviewing the specific solutions, it is necessary to firstly formalize the definitions of a grasp.
In this section, we will introduce several mainstream grasp representations, including the classical contact-based grasp representation, 6-D grasp representation, point-based grasp representation, oriented-rect grasp representation, and pixel-level grasp maps.
% The comparison of all presented representations is shown in \textcolor{red}{\figref{fig:grasp_repr}}.

\subsubsection{Contact-based Grasp Representation}
\label{sec:contact-grasp}

Formally, given a set of contact points $\pset=\{\p{i}\}_{i=1}^{\pnum}$, one wrench, $\w{i}=(f_i,\tau_i)$, is imposed on each contact point accordingly, where $f_i$ is the force exerted on the object at point $\p{i}$ and $\tau_i$ is the torque around the surface normal. A grasp $\g{j}$, $j\in\{1,2,...,\gnum\},$ is defined as $\g{j}=(\w{1}, \w{2}, ..., \w{\pnum})$, where all points, wrenches, and grasps are defined in the object reference frame. Obviously, if there is an external wrench, $\we$, imposed on the object, only when $\wg{j}=\sum_{i=1}^{\pnum}\w{i}=-\we$ can the object be in equilibrium.
This representation is widely used in analytic methods to be introduced in \secref{sec:analytic-grasp} and early data-driven approaches (e.g. \cite{kang1993toward, ekvall2005grasp, aleotti2006grasp}).
This kind of representation is scalable to different grippers with different numbers of fingers.
Therefore, it is still preferred in current days by grasping using dextrous hands.

\subsubsection{Independent Contact Regions}
\label{sec:icr-grasp}

Noticeably, the contact-based grasp representation is based on the ideal contact models, i.e., the contact points could be exactly positioned by the robot.
However, due to the inherent system or random errors, it would always be inaccurate to execute the planned grasps.
And certainly, such errors should be taken into consideration when synthesizing robust grasps.
Therefore, a more practical representation, named Independent Contact Regions (ICRs) \cite{nguyen1988constructing}, is introduced in spite of possibly introducing more computation.
It is defined as a set of independent regions on the object boundary such that putting one finger onto each ICR will result in a force-closure grasp (please refer to \secref{sec:analytic-grasp}) regardless of the exact position of each finger.

\subsubsection{$\sethree$ Grasp Representation}
\label{sec:se3-grasp}

% gripper pose
With the prevalence of parallel-jaw grippers, with some loss of scalability, it is more convenient to use simplified representations.
Since the kinematics of a parallel-jaw gripper is simple, the contact points on a specific object is completely determined by the gripper's 6-D pose $g=(x,y,z,r_x,r_y,r_z)\in\sethree$, including 3-D position $(x,y,z)$ and 3-D orientation $(r_x,r_y,r_z)$, which is a widely-used grasp representation based on 3-D perception \cite{miller2003automatic, gualtieri2016high, ten2017grasp, liang2019pointnetgpd}.
For the convenience of computation, $\sethree$ grasp representation may have different specific but equivalent forms in practice.

\subsubsection{Point-based Grasp Representation}
\label{sec:point-grasp}

% point based
Recently, as 2-D vision develops rapidly, it is feasible to directly synthesize grasps on RGB images instead of the 3-D point clouds, and hence, the grasp representation is further simplified. \cite{saxena2006robotic, saxena2008robotic} detected grasp points on multi-view observations and projected them back into a single 3-D grasp point. \cite{rao2010grasping} and \cite{asif2019densely} used segmented grasp affordance on 2-D images to represent grasps for parallel-jaw grippers. Such a single-dot representation is also widely used for suction grasps \cite{mahler2018dex, cao2021suctionnet, jiang2021learning}. Later, orientation was introduced to instruct the pose of the gripper \cite{mahler2017learning, wang2021double}.

\subsubsection{Oriented-rect Grasp Representation}
\label{sec:rect-grasp}

% oriented rectangles
The point-based grasp representation cannot model the size of the gripper. Moreover, it lacks a bounded feature area to map grasp points to robot configurations according to \cite{jiang2011efficient}. Therefore, they presented oriented rectangles for grasps on 2-D images. The oriented rectangle includes 5 dimensions: $g=(x,y,w,h,\theta)$, with $(x,y)$ denoting the center, $(w,h)$ denoting the distance between two jaws and the size of the gripper, and $\theta$ denoting the orientation of the gripper. It is now widely used in grasp detection with image inputs.

\subsubsection{Pixel-level Grasp Maps}
\label{sec:pixel-grasp-repr}

% pixel-wise grasp
Possible grasps are infinite on one object or one image. Therefore, based on the oriented-rectangle grasps, the pixel-level dense grasp representation was proposed \cite{asif2018graspnet, wang2019efficient, gkanatsios2020orientation, wang2021high}. Typically, it models the grasp synthesis as a segmentation problem. By taking as the input images, it outputs a segmented image of grasp affordance and possibly the corresponding gripper parameters. Formally, the grasp map $\gmap=({\bf Q}, {\bf W}, {\bf H},{\Theta})$ where ${\bf Q}$, ${\bf W}$, ${\bf H}$, and ${\Theta}$ are all single-channel images with the same size of the input, representing the pixel-wise graspability and the corresponding gripper parameters including width, height, and orientation. Note that not all elements in $\gmap$ are mandatory. For example, the height map ${\bf H}$ is not included in the output of the method in \cite{gkanatsios2020orientation}.

\section{Analytic Grasp Synthesis}
\label{sec:analytic-grasp}

\begin{table}[t]
\caption{Summary of Selected Analytic Grasp Synthesis}
\label{table:analytic-grasp}
\begin{center}
\begin{tabularx}{1\columnwidth}{m{4cm}m{3.5cm}<{\centering}m{1.8cm}<{\centering}m{1.3cm}<{\centering}m{3.3cm}<{\centering}}
\toprule
 \bf Author \& \bf Year & \bf Repr. & \bf Type & \bf Fingers & \bf Object \\
\midrule
\cite{nguyen1988constructing} & Contact points \& ICRs & Frictional & 2,3,4,7 & Polygons \& Polyhedra \\
\cite{markenscoff1989optimum} & Contact points & Frictionless & 3,4 & Polygons  \\
\cite{faverjon1991computing} & Contact points & Frictional & 2 & Curved Shapes \\
\cite{ferrari1992planning} & Contact points & Frictional & 2,3 & Polygons \\
\cite{ponce1995computing} & ICRs & Frictional & 3 & Polygons  \\
\cite{ponce1997computing} & ICRs & Frictional & 4 & Polyhedra \\
\cite{smith1999computing} & Contact points & Frictional & 2 & Polygons\\
\cite{liu2000computing} & Contact points & Frictional & n & Polygons \\
\cite{ding2001computation} & Contact points & Frictional & n & 3D Objects\\
\cite{zhu2003synthesis} & Contact points & Frictional & n & 3D Curved Objects\\
\cite{pollard2004closure} & Contact points \& ICRs & Frictional & n & 3D Objects \\
\cite{jia2004computation} & Contact points & Frictional & 2 &  Curved Shapes\\
% \cite{han2000grasp} & Contact points & Frictional & n & -  \\
\cite{cornelia2005determining} & ICRs & Frictionless & 4 & 2D Discrete Objects \\
\cite{cornella2005fast} & ICRs & Frictional & n & Polygons\\
\cite{niparnan2006computing} & Contact points & Frictional & 3 & 2D Objects \\
\cite{roa2008independent} & ICRs & Frictional & n & 3D Objects \\
\cite{roa2009computation} & ICRs & Frictional & n & 3D Objects \\

\bottomrule
\end{tabularx}
\end{center}
\vspace{-15pt}
\end{table}

In this section, we will review the mainstream approaches related to grasp synthesis.
By reviewing both classical and modern methods, we hope to inspire researchers to harness the best from both worlds for developing advanced grasping approaches.

\subsection{Overview}

% Definition
Analytic grasp synthesis is mostly based on mechanics.
Under this setting, a grasp is generally represented by a set of contact points and the corresponding wrenches imposed on each point \cite{bicchi2000robotic}.
% as shown in \textcolor{red}{Figure \ref{fig:grasp_analytic}}.
Typically, there are three different contact models:
\begin{itemize}
	\item {\bf Frictionless contact}, meaning that there is no friction at the contact point.
	\item {\bf Frictional contact}, meaning that there is friction at the contact point.
	\item {\bf Soft-finger contact}, meaning that the contact part is deformable and will be an area instead of a point, and thus allows an additional torque around the surface normal.
\end{itemize}
We focus on the first two types of contact models, which are most widely explored in robotics.
For the soft-finger contact models, we refer to \cite{elango2015review} for more detailed discussions.

In detail, we firstly review the methods to analytically evaluate the quality of a given grasp.
With a given evaluation metric, it would be non-trivial to synthesize the optimal grasps. Typically, either heuristic or analytic methods could be applied to the computation and optimization of grasps. A summary of the included analytic grasp synthesis methods is demonstrated in \tabref{table:analytic-grasp}.

\subsection{Grasp Quality Evaluation}
\label{sec:analytic-grasp-eval}

Form-closure and force-closure are often used to evaluate the quality of a given grasp provided the frictionless and frictional models respectively \cite{bicchi1995closure}.
It is often assumed that the object models are fully observable, including the geometry and the friction coefficient at each contact point.
One can regard the form-closure grasp synthesis as a special case of force-closure with frictionless contact points.
They are both defined as follows:
\begin{definition}
\label{def:grasp-closure}
	{\bf (Grasp-closure)} is a property of a given grasp, including {\bf form-closure} for grasps with frictionless contact points and {\bf force-closure} for grasps with frictional contact points. It occurs only when the grasp could resist any possible external disturbing wrenches. 
\end{definition}

The research on grasp-closure can trace back to 19$^{th}$ century \cite{reuleax1963kinematics}. They proved that for a 2-D polygon, at least 4 frictionless wrenches are required for form-closure grasping. 
Much later, \cite{lakshiminarayana1978mechanics} showed that at least 7 contact points are needed for 3-D polygons.
Based on their analysis, \cite{markenscoff1990geometry} proved their conjectures.
In particular, they showed that iff without rotational symmetry, form-closure for any 3-D bounded object with piecewise smooth boundary could be achieved by 12 fingers, and in most cases, 7 fingers could be enough.
Moreover, they also demonstrated that with Coulomb friction, the required number of fingers to achieve force-closure could reduce to 3 and 4 for 2-D and 3-D objects respectively under certain circumstances.
Later, the definitions of form-closure and force-closure were formally completed by \cite{bicchi1995closure}.
Following that, \cite{rimon1996force} argued that the previous definition for form-closure and force-closure (referred by 1$^{st}$ grasp-closure) are not adequate and should be considered together with the mobility of the fingers. 
They proposed the definition of 2$^{rd}$ form-closure and force-closure to fix the deficiency.

Though the grasp-closure property has been well-investigated, one could notice that the definition in \defref{def:grasp-closure} is unduly strict.
In practice, the forces that the actuator could impose on the object are usually limited.
Therefore, a more practical metric is needed to evaluate a given grasp.
One natural way to evaluate a grasp is the minimum force needed to achieve equilibrium \cite{markenscoff1989optimum, pollard2004closure, liu2004quality}, and the directions of imposed forces should be close to the surface normals for stability \cite{han2000grasp, liu2004quality}.
However, such methods usually assume that the accessibility of the externally exerted wrenches on the object.

To improve this deficiency, \cite{ferrari1992planning} proposed to utilize the Grasp Wrench Space (GWS) to evaluate the quality of grasps:
\begin{definition}
\label{def:gws}
	{\bf (Grasp Wrench Space)} of grasp $\g{i}$ is defined as the convex hull of all possible wrenches that could be imposed through the contact points $\{\p{i}\}_{i=1}^{\pnum}$ of grasp $\g{i}$.
\end{definition}
In particular, the minimum distance between the origin and the boundary of the GWS, called the Largest-minimum Resisted Wrench (LRW), represents the minimum external wrench that could affect the stability of the object.
It quickly became one of the most well-known metrics for grasp quality evaluation.
Based on GWS and LRW, \cite{mishra1995grasp} proposed that different distances such as $L_1$ and $L_\infty$ could be applied to the measurement of LRW.
\cite{mirtich1994easily} decoupled the forces and torques of wrenches in the wrench space to avoid the balancing factor between them.
\cite{teichmann1996grasp} and \cite{miller1999examples} proposed to use the volume of GWS instead of the distance to get rid of the dependence on the predefined reference frame on the object.

There are also other metrics used for quality evaluation in analytic grasp synthesis, such as 
the shape \cite{park1992grasp, kim2001optimal} and volume \cite{mirtich1994easily, chinellato2003ranking, supuk2005estimation} of the grasp polygon formed through all contact points, the distance between the centers of the object and the grasp polygon \cite{ponce1997computing, ding2001computation, chinellato2005visual}, and the size or radius of ICRs \cite{stam1992system, ponce1995computing, cornelia2005determining}.
We refer to the review by \cite{roa2015grasp} for more detailed discussions.

\subsection{Grasp Synthesis on Simple Shapes}
\label{sec:hard-grasp-simple}

Early works mostly focused on the grasp synthesis for simple shapes like polygons or polyhedra, which approximately satisfy the assumptions made by grasp closure properties.
\cite{nguyen1988constructing} developed the principles for the force-closure grasp synthesis. They proposed three basic contact types: frictionless contact, hard-finger (frictional) contact, and soft-finger contact, and proposed that any complex contact types like edge contact and face contact could be factorized using the three basic types.
Based on their analysis, they also developed methods for finding the force-closure grasps and ICRs for simple polygons and polyhedra.
\cite{markenscoff1989optimum} applied elementary optimization techniques to the synthesis of grasps for any polygons with the minimization of the needed forces to balance the object.
\cite{ferrari1992planning} proposed GWS for grasp evaluation along with an iterative heuristic method for searching the optimal grasp on polygons with two- or three-finger grippers.
\cite{ponce1995computing} proved new sufficient conditions for force-closure grasping of polygons, resulting in a more efficient polygonal grasping method with linear optimization.
Later, they developed the method to handle 3-D polyhedra with a four-finger gripper \cite{ponce1997computing}.
\cite{smith1999computing} simplified 3-D objects by their intersections, and planned two-point grasps on the corresponding polygon for parallel-jaw grippers considering the uncertainty of the mass center.
\cite{liu2000computing} presented new sufficient and necessary conditions for form-closure grasping of polygons with $n$-finger grippers.
\cite{han2000grasp} formalized the {\it force closure}, {\it force feasibility}, and {\it force optimization} problems in a unified convex optimization problem with linear matrix inequalities, and solved it numerically in polynomial time.
\cite{cornella2005fast} proposed a fast approach based on the two-dimensional problem formulation in the object space instead of the contact space to efficiently synthesize ICRs.

Though theoretically sound and optimal, such kinds of methods always rely on the simplification on contact models and geometry of objects, which severely restricts their application in real-world scenarios.

\subsection{Grasp Synthesis on General Shapes}

Following the methods mentioned in \secref{sec:hard-grasp-simple}, researchers explored the ways to relax the shape assumption to improve the real-world grasping performance. 
\cite{faverjon1991computing} focused on the force-closure grasp synthesis for parametrically curved objects instead of simple polygons.
Later, \cite{jia2004computation} proposed the method to compute all pairs of antipodal points \cite{chen1993finding} on twice continuously differentiable shapes.
\cite{ding2001computation} derived the sufficient and necessary conditions of an incremental method to construct a $n$-finger force-closure grasp (though referred by form-closure in their paper) given a $k$-finger non-force-closure grasp for any 3-D objects.
\cite{zhu2003synthesis} presented the Q-distance and applied it to the computation of force-closure grasps. With the differentiable Q-distance, they could apply gradient descent to find the optimal grasp configuration.
\cite{pollard2004closure} argued that more contact points could result in more stable grasps.
Therefore, they proposed a method to handle a large number of contacts in polynomial time w.r.t. the contact number instead of exponential time previously on any object geometry.
Their method searched the ICRs efficiently based on initial examples.
\cite{cornelia2005determining} considered the uncertainty of the object description and proposed to compute the ICRs without the iterative search.
\cite{niparnan2006computing} illustrated that it is feasible to achieve a complexity of $O(n^2\log^2 n+K)$ to get $K$ three-finger solutions from $n$ candidates of contract points on an arbitrary shaped 2-D object.
Similar to \cite{pollard2004closure}, a method based on the initial examples is proposed to incrementally synthesize ICRs on arbitrary 3-D objects in \cite{roa2008independent}.
Following their previous work, they generalized their method to the computation of ICRs with any number and type of contact points with hard-finger grippers \cite{roa2009computation}.

Methods presented in this section relax the shape assumptions and are certainly more flexible and practical. However, there are still some assumptions preventing them from the widespread application. For example, some methods assume that the objects could be represented by parametric curves \cite{faverjon1991computing, zhu2003synthesis, jia2004computation}, and all the methods require the complete 2-D or 3-D model of objects including the friction coefficients, which do not always hold in real-world scenarios. Though some works tried to get rid of the impractical assumptions by the online estimation of object models \cite{kragic2001real, rosales2012synthesis, rodriguez2012caging, zhang2012application}, such methods are always not satisfactory due to the gap between the models and reality.

%\paragraph{Soft-finger Grasp Synthesis}
%
%For the soft contact model, it would be more complicated and could only be analytically solved in some special cases according to \cite{ciocarlie2005grasp}. They also provide a numerical solution based on finite element modeling (FEM) to handle this complex contact type.

\section{Data-driven Grasp Synthesis}
\label{sec:data-driven-grasp}
% Definition
As machine learning technology develops rapidly, it is promising to learn robot skills by training on a large amount of data instead of planning with object models \cite{bohg2013data, kroemer2021review}.
Such methods are always called data-driven since the quality and quantity of data are also essential parts for a good policy besides the methods.
In this section, we will review a series of data-driven grasp synthesis approaches.

\subsection{Overview}

% comparison with traditional visual tasks and analytic methods.
In most cases, modern grasp synthesis is based on perception, especially the visual observations of the workspace.
However, different from traditional visual tasks, grasp synthesis usually involves the precise perception and analysis of geometric information, and sometimes intuitive physics, especially when facing unknown objects.
And compared to analytic methods, data-driven approaches substantially loosen the assumptions of accessible object models since inspired by the neuropsychology \cite{jeannerod1988neural}, it is widely found that the heuristic abstraction of knowledge is enough to derive reliable robotic control signals \cite{allen1985object, stansfield1987visually, iberall1988knowledge, liu1989multi, Nayar1994, salganicoff1996active, kamon1996learning, Piater2001}.

% taxonomy
As its name implies, the provided data plays a role of ``experiences'', driving the robot to abstract ``knowledge'' adaptively for skill learning.
Different methods implement the abstraction in different ways.
Regarding robotic grasping, there are mainly three ways:
\begin{itemize}
	\item {\bf Imitation-based methods}: Given a dataset including stable grasps (e.g. force-closure grasps) and the corresponding objects, the grasps could be transferred to similar objects by imitation. The imitative policy could be formulated through the similarity between the target and object templates in the given dataset, or between the real robot configuration and the given grasp templates. Early works focused more on this type of method since it is more data-efficient.
	\item {\bf Sampling-based methods}: To generate grasps on objects, another way is to sample a set of possible candidates, among which a discriminator is used to find the best one. Benefiting from the decoupling of grasp sampling and classification, it has better interpretability and scalability. Nevertheless, it relies heavily on a better grasp sampler in terms of both performance and running speed.
	\item {\bf End-to-end Learning}: With the development of deep learning, it is possible to embed all things into one neural network model and train it end-to-end. The input could be the raw observations such as information from tactile sensors or cameras. And the output is proper grasp configurations. All steps including grasp sampling and quality evaluation could be adaptively tuned with updates of trainable parameters. Such methods usually run faster than the above two types and benefit mostly from large datasets.
% TODO: reinforcement learning
\end{itemize}

\subsection{Imitation-based Methods}

\begin{table}[t]
\caption{Summary of Selected Imitation-based Grasp Synthesis}
\label{table:imitation-grasp}
\begin{center}
\begin{tabularx}{1\columnwidth}{m{5cm}m{1.8cm}<{\centering}m{2.4cm}<{\centering}m{2.2cm}<{\centering}m{2.6cm}<{\centering}}
\toprule
 \bf Author \& \bf Year & \bf Imit. Type & \bf Modality & \bf Abstractor & \bf Planner \\
\midrule

\cite{kang1993toward} & PbD & Vision + Tactile & Heuristics & K. Map \\
\cite{kragic2001real} & MOOT & Vision & Pose Est. & GraspIt!\\
\cite{zollner2001dynamic} & PbD & Vision + Tactile & SVM & K. Map\\
% \cite{miller2003automatic} & MOST & Vision & - & GraspIt!\\
\cite{ekvall2005grasp} & PbD & Trajectories & HMM + Similarity & -\\
\cite{aleotti2006grasp} & PbD & Trajectories & Similarity & K. Map \\
\cite{hsiao2006imitation} & MOOT & Vision & Similarity & G. Map\\
\cite{morales2006anthropomorphic}  & MOOT & Vision & Similarity & G. Map\\
\cite{ekvall2007learning} & PbD + MOST & Vision + Tactile & HMM + Similarity & K. Map + GraspIt! \\
\cite{sweeney2007model} & Generative & Vision & - & PGM \\
\cite{curtis2008efficient} & MOOT & Vision & Similarity & G. Map \\
\cite{do2009grasp} & PbD & 2D Vision & Similarity & K. Map\\
\cite{grave2010improving} & PbD & Trajectories & GPR & K. Map + RL\\
\cite{schmidts2011imitation} & PbD & Trajectories & HMM & GMR \\
\cite{herzog2012template} & MOOT & Vision & Similarity & G. Map\\
\cite{pokorny2013grasp} & MOST & Vision & Similarity & Interpolation\\
\cite{lin2014grasp} & PbD & Trajectories & Similarity & K. Map + Search \\

\bottomrule
\end{tabularx}
\end{center}
\vspace{-15pt}
\end{table}

Early works often relied on imitation-based methods to extract knowledge from the given data.
Two ideas were often considered as solutions: 1) programming by demonstrations (PbD); 2) matching of templates (MoT).
We will discuss them respectively.
A selected set of imitation-based grasp synthesis methods are also summarized in \tabref{table:imitation-grasp}.

\subsubsection{Programming by Demonstrations}
\label{sec:pbd}

PbD means that successful grasping trajectories are recorded first.
When testing, the robot will adjust and replay the trajectory to grasp objects.
Grasp recognition is one crucial component in PbD-based grasp synthesis and was widely investigated \cite{kang1993toward, zollner2001dynamic, ekvall2005grasp, aleotti2006grasp, de2006learning}.
It assigns a specific category to a given grasp configuration from a predefined taxonomy \cite{cutkosky1989grasp, zollner2001dynamic, feix2009comprehensive}.
Based on the recognized grasp type and the demonstration, the planner could synthesize the grasp for the robot using mapping of kinematics \cite{aleotti2006grasp}, or search efficiently in a constrained grasp space \cite{lin2014grasp, lin2015robot}.
The demonstration could be also combined with reinforcement learning to incrementally improve the performance, achieving better adaptation to the robot \cite{grave2010improving}.

Generative models are also feasible for PbD-based grasp synthesis.
Demonstrations are used to train the generator.
In the phase of testing, the trained generator takes as input features of objects, and output a distribution from which one can sample grasps \cite{sweeney2007model, schmidts2011imitation, arruda2019generative}.

As machine learning develops, behavior cloning \cite{zhang2018deep} and inverse reinforcement learning \cite{abbeel2004apprenticeship, horn2017quantifying, xie2019learning} have also been explored in the context of robotic grasping.
Behavior cloning transforms imitation learning to supervised learning, in which the demonstrations are regarded as a labeled dataset, based on which a model is trained to map from inputs to actions.
Inverse reinforcement learning is used to infer an explanative reward for the given demonstrations. The inferred reward is then applied to policy training with reinforcement learning.

Recently, meta-learning \cite{vanschoren2018meta} enables few-shot and even one-shot imitation directly from raw visual observations such as videos or images.
Meta-learning is the basis for one-shot imitation learning \cite{duan2017one, finn2017one, mandi2021towards}, in which a model is trained using some off-the-shelf data first to get a meta policy, and during imitation learning, it will be fine-tuned by the demonstration to get the final policy.
It is even possible to transfer the given demonstration across different agents with different kinds of morphology \cite{yu2018one, yu2018oneb, dasari2020transformers, bonardi2020learning}
Besides, data augmenting is also an interesting idea to achieve few-shot imitation learning \cite{de2019learning}.
The provided demonstrations will be augmented by a delicately designed pipeline and used to train a neural network.
However, the achieved generality is limited compared to meta-learning methods. 

\subsubsection{Matching of Templates}
\label{sec:template-match}

MoT can be classified into two subclasses: 1) Matching of Object Template (MOOT) and 2) Matching of Shape Template (MOST).

In MOOT, a set of object templates along with their corresponding grasps are usually predefined.
When the robot meets novel objects, it will look up the template set and find the most similar one so as to map the predefined grasps onto targets.
One straightforward way of doing so is by manipulation-oriented pose estimation \cite{kragic2002model, ekvall2003object, collet2009object, marchand2015pose, tremblay2018deep}.
Concretely, objects will be recognized and positioned first, and predefined grasps from demonstrations could be directly projected into the reference frame of objects for grasping \cite{morales2006anthropomorphic, do2009grasp}, or simulation-based grasp planners could be introduced to online grasp planning \cite{kragic2001real, miller2004graspit, berenson2009manipulation}.
However, such methods could be only used for known objects, i.e., it is usually assumed 3-D object geometric models (e.g. the mesh or point cloud) are available for estimation of 6-D poses.

To grasp unknown objects, shape primitives were proposed, resulting in MOST.
Instead of full objects, they form a set of primitive shapes, based on which grasps are predefined.
The primitive set could be infinite \cite{pokorny2013grasp}.
When grasping novel objects, a matching process will be conducted between the predefined primitives and the target.
Then the demonstrated grasps could be mapped and executed \cite{miller2003automatic, hsiao2006imitation, ekvall2007learning, curtis2008efficient, herzog2012template, herzog2014learning} or a grasp planner could be incorporated based on the matched primitives \cite{goldfeder2007grasp, huebner2008minimum, el2008handling}.
Such methods could grasp objects with similar appearance to the primitives, but cannot handle objects with unknown geometric structures.

\subsection{Sampling-based Methods}
\label{sec:sample-grasp}

\begin{table}[t]
\caption{Summary of Selected Sampling-based Grasp Synthesis}
\label{table:sampling-grasp}
\begin{center}
\begin{tabularx}{1\columnwidth}{m{4cm}m{1.3cm}<{\centering}m{2cm}<{\centering}m{2cm}<{\centering}m{2.7cm}<{\centering}m{1.6cm}<{\centering}}
\toprule
 \bf Author \& \bf Year & \bf Repr. & \bf Modality & \bf Generator & \bf Discriminator & \bf View\\
\midrule

\cite{saxena2006robotic} & Point & RGB & SW & LR & Multi \\
\cite{goldfeder2007grasp} & Contacts & 3D Models & Heuristics & GraspIt! & - \\
\cite{huebner2008minimum} & Contacts & 3D Models & Heuristics & GraspIt! & - \\
\cite{bohg2009grasping} & Point & RGB & Seg + SW & SVM & Two\\
\cite{rao2010grasping} & Point & RGB & Seg & SVM & Single \\
%\cite{le2010learning} & \\
\cite{bekiroglu2011assessing} & Contacts & Tactile & - & SVM, AdaBoost, HMM & - \\
\cite{jiang2011efficient} & Rect & RGB-D & SW & SVM & Single\\
\cite{dang2012learning} & Contacts & Tactile & - & SVM & -\\
\cite{schill2012learning} & Contacts & Tactile & - & SVM & -\\
\cite{lenz2015deep} & Rect & RGB-D & NN & NN & Single \\
\cite{gualtieri2016high} & SE(3) & Point Clouds & Random & CNN & Single\\
\cite{mahler2017dex} & Rect & Depth & CEM & CNN & Single \\
\cite{mahler2017learning} & Rect & Depth & CEM & CNN & Single\\
\cite{mahler2018dex} & Point & Depth & CEM & CNN & Single \\
\cite{ten2018using} & SE(3) & Point Clouds & Heuristics & SVM & Single\\
\cite{liang2019pointnetgpd} & SE(3) & Point Clouds & Heuristics & NN & Single\\
\cite{mousavian20196} & SE(3) & Point Clouds & VAE & NN & Single\\
\cite{yan2019data} & SE(3) & Point Clouds & CEM & NN & Single\\
\cite{li2020learning} & SE(3) & Point Clouds & CNN & Heuristics & Single \\
\cite{chen2021ab} & Rect & RGB & CEM & Physical Simulator & Multi \\
\cite{gou2021rgb} & SE(3) & RGB-D & CNN & Heuristics & Single \\

\bottomrule
\end{tabularx}
\end{center}
\vspace{-15pt}
\end{table}

Sampling-based methods derived from the extensively explored grasp quality evaluation approaches, and hence, are widely applied to grasp synthesis with 3-D perception.
The main idea is to select the best grasp according to a well-trained discriminator provided a set of sampled candidates.
Therefore, two main components are equally important: 1) the discriminator; 2) the sampler.
Noticeably, different from analytic grasp quality evaluation introduced in \secref{sec:analytic-grasp-eval}, the discriminator is designed to get rid of the strong reliance on object models by making use of large-scale datasets and learning techniques.
The included algorithms are summarized in \tabref{table:sampling-grasp}.

\subsubsection{Discriminator}

To learn the discriminator, supervised learning is usually applied.
Noticeably, learning of discriminators is similar to grasp recognition of PbD (\secref{sec:pbd}): both of them are trained using labeled data and supervised learning.
The essential difference is that in grasp recognition of PbD, the grasp category based on the predefined taxonomy is the focus, which is important to decide a certain grasp pattern for the robot.
By contrast, the discriminator here is used to evaluate whether a grasp is good or not.
Possible quality metrics could either derive from analytic methods \cite{roa2015grasp}, or a simple indicator of success or failure \cite{saxena2006robotic, saxena2008robotic}.

\begin{table}[t]
\caption{Summary of Selected Robotic Grasp Dataset}
\label{table:dataset}
\begin{center}
\begin{tabularx}{1\columnwidth}{m{5.5cm}m{2cm}<{\centering}m{1.5cm}<{\centering}m{1cm}<{\centering}m{1cm}<{\centering}m{1.1cm}<{\centering}m{1cm}<{\centering}}
\toprule
%                             2D 3D Multi                SL SSL RL    one-stage two-stage
 \bf Dataset & \bf Repr. & \bf Modality & \bf Source & \bf Size &\bf Object /Scene & \bf Grasp /Scene \\
\midrule
Cornell \cite{jiang2011efficient}  & Rect & RGB-D & Real & 1035 & 1 & $\sim$8 \\
Dex-Net 2.0 \cite{mahler2017dex} & Rect & Depth & Sim & 6.7M & 1 & 1 \\
Dex-Net 3.0 \cite{mahler2018dex} & Point & Depth & Sim & 2.8M & 1 & 1 \\
Jacquard \cite{depierre2018jacquard} & Rect & RGB-D & Sim & 54K & 1 & $\sim$20 \\
VMRD \cite{zhang2018visual} & Rect & RGB & Real & 4.7K & $\sim$3 & $\sim$20\\
\cite{levine2018learning} & - & RGB-D & Real & 800K & - & 1\\
\cite{wang2019multimodal} & - & RGB-D-T & Real & 2.55K & 1 & 1 \\
GraspNet-1billion \cite{fang2020graspnet} & Rect + SE(3) & RGB-D & Real & 97K & $\sim$10 & 3-9M\\
ACRONYM \cite{eppner2021acronym} & SE(3) & Depth & Sim & 8.8K & 1 & 2K \\
SuctionNet-1billion \cite{cao2021suctionnet} & Point & RGB-D & Real & 97K & $\sim$10 & 3-8M\\
REGRAD \cite{zhang2022regrad} & Rect + SE(3) & RGB-D & Sim & 900k & 1-20 & 1.02K\\
\bottomrule
\end{tabularx}
\end{center}
\end{table}

Before the prevalence of deep learning, Support Vector Machines (SVM) \cite{cortes1995support} or probabilistic models \cite{koller2009probabilistic} are widely used to train such a discriminator \cite{pelossof2004svm, saxena2006robotic, saxena2008robotic, bohg2009grasping, le2010learning, bohg2010learning, bekiroglu2011assessing, jiang2011efficient, schill2012learning, dang2012learning}.
Training data are either collected in the real world and manually labeled \cite{jiang2011efficient, schill2012learning} or automatically synthesized using physical simulators \cite{pelossof2004svm, saxena2006robotic, saxena2008robotic, bekiroglu2011assessing, goldfeder2009columbia}.
However, if synthetic data are used, there might be a reality gap when trained models are applied in real-world scenarios due to domain shift \cite{wang2018deep}.
Sizes of datasets in this period are always limited since such models are quite data-efficient and could achieve commendable performance with a few (usually hundreds of) data points.

As deep learning shows categorical advantages over other methods, it dominates learning of grasp discriminators recently.
Nevertheless, compared to SVM, it needs much more data to train a good model.
Therefore, datasets including more and more data are proposed to meet the demands of deep networks \cite{bullock2015yale, calli2017yale, mahler2016dex, mahler2018dex, depierre2018jacquard, wang2019multimodal, fang2020graspnet, eppner2021acronym, cao2021suctionnet, zhang2018visual, zhang2022regrad}.
A summary of robotic grasp datasets is shown in \tabref{table:dataset}.
One also could refer to \cite{huang2016recent} for a comprehensive summary of large-scale robotic manipulation datasets.
Generally speaking, deep-learning-based grasp discriminators are in essence the same as SVM-based discriminators despite much stronger representability and performance.
The main difference is that deep networks support much complex input data modality, such as raw 2-D images \cite{lenz2015deep} and point clouds \cite{gualtieri2016high, ten2017grasp, liang2019pointnetgpd, li2020learning}.

There are also methods not relying on the learned discriminators to evaluate the quality of grasps.
In this case, a model of the target is usually needed to be estimated first, such as the 3-D shape, friction, and center of mass.
Such a model is not necessarily accurate in most cases.
For example, PROMPT \cite{chen2021ab} only builds a 3-D particle-based model through multi-view images for the target and applied NVIDIA Flex with predefined friction to the evaluation of whether a grasp will succeed or not.
It chooses the best sample for real-world execution.
By comparison of the difference between the simulator and the reality, PROMPT could update parameters of object models in an online and close-loop way.
Many grasp planners based on physical properties such as \cite{kragic2001real}, \cite{miller2004graspit}, and \cite{berenson2009manipulation} are possible to be introduced here to replace learning-based discriminators given the estimation of object models.
For example, GraspIt! \cite{miller2004graspit} is widely used in the early works for grasp quality evaluation given a reasonable set of grasp candidates from learned samplers.

\subsubsection{Sampler}

A sampler could be either data-driven or heuristic.
Different data modalities and grasp representations usually correspond to different sampling methods.

For image inputs, the sampler is used to sample points for point-based grasp representation (\secref{sec:point-grasp}), or image patches for oriented-rect grasp representation (\secref{sec:rect-grasp}). 
One naive way to sample points is pixel-wise random sampling.
However, it is inefficient and sometimes intractable because: 1) sample space is too large; 2) a point is not representative and does not include enough features to indicate the quality of a grasp.
Therefore, learning is used to obtain a prior for sampling.
In this case, the output of a learning-based sample is usually a grasping affordance map, with higher values denoting the more graspable area \cite{saxena2006robotic, saxena2008robotic, bohg2009grasping, rao2010grasping, bohg2010learning, li2020learning, gou2021rgb}.
Based on the affordance map, all points could be ranked and tested one by one to find the best grasp configuration.
To solve the problem of representability, the sampled point could be mapped to 3-D space with camera models \cite{saxena2006robotic, saxena2008robotic, bohg2009grasping, bohg2010learning, gou2021rgb}, or extended to a full grasp configuration heuristically \cite{rao2010grasping}.
To sample image patches for oriented-rect grasp synthesis, random sampling methods such as the sliding window (SW) method are intractable due to unacceptable time complexity.
A more efficient way is to learn a patch sampler by taking the image as the input as long as the inference speed of the learned sampler is much faster than the discriminator \cite{jiang2011efficient, lenz2015deep, wang2016robot}.
Moreover, some heuristics could be used to further reduce search space.
For example, given a predefined patch size and the assumption that the background is a flat table, one could uniformly sample surface normals computed from depth gradients \cite{mahler2017dex, mahler2018dex}.

For point cloud inputs, the sampler is used to sample different $\sethree$ grasp poses (\secref{sec:se3-grasp}) in most cases.
Different from 2-D points, 3-D points include much richer geometric information which will help to efficiently filter out undesired regions.
For example, \cite{gualtieri2016high} and \cite{ten2017grasp} voxelized and uniformly sampled points in regions of interest, and performed local grid search to generate a set of $\sethree$ grasp candidates.
Finally, they filtered out the ones causing collisions between the gripper or including no object points within the closing region of the gripper.
\cite{ten2018using} improved grid search in this sampling method for higher efficiency, which is applied and further modified by \cite{liang2019pointnetgpd}.
An alternative way is to apply the Cross Entropy Method (CEM) \cite{rubinstein2004cross} starting from a randomly sampled grasp set and finally converging to an optimal graspable point distribution \cite{mahler2017dex, mahler2018dex, yan2019data}.
Learning-based samplers are also feasible and show higher efficiency especially in terms of speed \cite{mousavian20196, yang2021robotic}.
For antipodal grasps \cite{chen1993finding}, a mapping between single points and grasps could be built on object meshes, which results in a simplification from grasp sampling to point sampling \cite{rao2010grasping, mahler2017dex}.

\subsection{End-to-end Learning}
\label{sec:e2e-grasp}

\begin{table}[t]
\caption{Summary of Selected End-to-end Grasp Synthesis}
\label{table:e2e-grasp}
\begin{center}
\begin{tabularx}{1\columnwidth}{m{3.5cm}m{2cm}<{\centering}m{2cm}<{\centering}m{1.6cm}<{\centering}m{1.6cm}<{\centering}m{1.2cm}<{\centering}m{1.2cm}<{\centering}}
\toprule
%                             2D 3D Multi                SL SSL RL    one-stage two-stage
 \bf Author \& \bf Year & \bf Repr. & \bf Modality & \bf Method  &\bf Structure & \bf Anchor &\bf Gripper \\
\midrule
\cite{redmon2015real} & Rect & RGB-D & Open-loop & 1-stage & \xmark & Parallel \\
\cite{guo2016deep} & Rect & RGB & Open-loop & 1-stage & Vertical & Parallel \\
\cite{guo2017hybrid} & Rect & RGB-T & Open-loop & 1-stage & Vertical & Parallel \\
\cite{kumra2017robotic} & Rect & RGB-D & Open-loop & 1-stage & \xmark & Parallel \\
\cite{chu2018real} & Rect & RGB-D & Open-loop & 2-stage & Vertical & Parallel\\
\cite{levine2018learning} & - & RGB & Close-loop & 1-stage & \xmark & Parallel \\
\cite{morrison2018closing} & GMap & Depth & Close-loop & 1-stage & \xmark & Parallel \\
\cite{zeng2018learning} & GMap & RGB-D & Close-loop & 1-stage & \xmark & Parallel \\
\cite{zhou2018fully} & Rect & RGB & Open-loop & 1-stage & Oriented & Parallel \\
\cite{asif2019densely} & Rect + GMap & RGB-D & Open-loop & 1-stage & \xmark & Parallel \\
\cite{gariepy2019gq} & Rect & Depth & Open-loop & 4-stage & \xmark & Parallel \\
\cite{liu2019active} & GMap & RGB-D & Close-loop & 1-stage & \xmark & Parallel \\
\cite{morrison2019multi} & GMap & Depth & Close-loop & 1-stage & \xmark & Parallel\\
\cite{shao2019combining} & GMap & RGB-D & Open-loop & 1-stage & \xmark & Suction \\
\cite{shao2019suction} & GMap & RGB-D & Open-loop & 1-stage & \xmark & Suction\\
\cite{chalvatzaki2020orientation} & GMap & RGB-D & Open-loop & 1-stage & \xmark & Parallel\\
\cite{zhang2019real} & Rect & RGB-D & Open-loop & 1-stage & Oriented & Parallel \\
\cite{ni2020pointnet++} & SE(3) & Point Clouds & Open-loop & 1-stage & \xmark & Parallel \\
\cite{qin2020s4g} & SE(3) & Point Clouds & Open-loop & 1-stage & \xmark & Parallel \\
\cite{shao2020unigrasp} & SE(3) & Point Clouds & Open-loop & $n$-stage & \xmark & Dextrous \\
\cite{wu2020grasp} & SE(3) & Point Clouds & Open-loop & 1-stage & Point & Parallel \\
\cite{cao2021suctionnet} & GMap & RGB-D & Open-loop & 1-stage & \xmark & Suction \\
\cite{sundermeyer2021contact} & SE(3) & Point Clouds & Open-loop & 1-stage & \xmark & Parallel \\
\cite{wang2021graspness} & SE(3) & Point Clouds & Open-loop & 2-stage & \xmark & Parallel \\
\cite{wei2021gpr} & SE(3) & Point Clouds & Open-loop & 2-stage & \xmark & Parallel \\
\cite{wang2021double} & Point & RGB-D & Open-loop & 1-stage & \xmark & Parallel \\
\cite{wu2021real} & Rect & RGB-D & Open-loop & 1-stage & \xmark & Parallel \\
\cite{xu2021gknet} & Point & RGB-D & Open-loop & 1-stage & \xmark & Parallel \\
\cite{xu2021adagrasp} & GMap & RGB-D & Open-loop & 1-stage & \xmark & Dextrous \\
\cite{zhao2021regnet} & SE(3) & Point Clouds & Open-loop & 3-stage & SE(3) & Parallel \\

\bottomrule
\end{tabularx}
\end{center}
\vspace{-15pt}
\end{table}

End-to-end learning of grasp synthesis means that a model will be trained taking as the input raw observations (e.g. RGB images or point clouds), and directly output the best grasp to be executed.
There are mainly two ideas to do so: 1) grasp detection inspired by object detection; 2) pixel-level grasp map synthesis inspired by scene segmentation.
We will review both of them in this section.
A summary is also available in \tabref{table:e2e-grasp}.

\subsubsection{Grasp Detection on Images}

Deep convolutional networks enable end-to-end visual perception based on image inputs \cite{krizhevsky2012imagenet, simonyan2014very, he2016deep}.
To detect grasps on images, the most straightforward way is to directly transfer object detection algorithms to the domain of grasp detection, since detection algorithms share a similar basis: they both are essentially a classification problem based on a set of extracted proposals.
From this view, end-to-end learning also shares a similar idea as sampling-based methods.
The difference is that end-to-end learning integrates the sampler and the discriminator into one single model and trains them end-to-end.
For example, \cite{guo2016deep, chu2018real} transfers Faster R-CNN \cite{ren2015faster} to a two-stage grasp detection algorithm.
And vise versa, grasp detection like \cite{redmon2015real} sometimes also inspired object detection research \cite{redmon2016you}.

Nevertheless, grasp detection is essentially different from object detection since 1) grasp detection relies heavily on the local geometry of grasps; 2) grasp quality is sensitive to orientations; 3) grasping should be a close-loop process, meaning that failures should be handled during grasping.
For 1), image-based grasp detection algorithms usually take a combination of color and geometric channels, such as depth images as input \cite{redmon2015real, kumra2017robotic, chu2018real, zhang2019real, song2020novel, wu2021real} and surface normals \cite{park2018classification}.
For 2), the dimension of the orientation could be discretized and the prediction of orientations could be simplified as a classification problem \cite{chu2018real, wu2021real}.
However, discretization suffers from performance loss, especially for orientation-sensitive grasps.
Therefore, oriented anchors were introduced to handle this problem \cite{zhou2018fully, zhang2019real}.
Besides, Spatial Transformer Network (STN) \cite{jaderberg2015spatial} could also be used for more accurate classification of oriented grasp candidates \cite{park2018classification, gariepy2019gq}.
Recently, \cite{park2020real} proposed rotation ensemble module to handle rotation-invariance for grasp detection.
For 3), reactive policies could be trained for grasping by taking raw images as inputs \cite{baier2007learning, lampe2013acquiring, mahler2017learning, levine2018learning, zeng2018learning}.
% quillen2018deep, kalashnikov2018scalable, breyer2018flexible, gualtieri2018learning, deng2019deep, pedersen2020grasping, shahid2020learning, chen2020towards

% TODO: discuss the reactive methods

\subsubsection{Grasp Detection on Point Clouds}

Recent developments in 3-D vision enables end-to-end learning with point clouds as inputs \cite{qi2017pointnet, qi2017pointnet++, li2018pointcnn, zhou2018voxelnet, zhao2021point, mazur2021cloud}.
Such methods have also been explored for grasp synthesis.
In this case, a backbone is usually used to pre-process input point clouds, including subsampling, grouping, de-noising, etc., following which a feature extractor designed based on strong inductive bias is used to extract features of points.
The extracted features are then fed into a grasp detector to regress $\sethree$ grasps as well as confidence scores indicating grasp quality of each point correspondingly.
The most simple framework is the one-stage anchor-free grasp detection \cite{qin2020s4g, ni2020pointnet++, sundermeyer2021contact}, which directly output results right after the feature extraction stage.
To improve performance, $\sethree$ grasp anchors and sphere-region features instead of point features were introduced in \cite{zhao2021regnet}.
Their method also includes a fine-tune stage, which further improves robustness.
A similar idea has also been explored by \cite{wu2020grasp, wei2021gpr}.
Another problem is that when synthesizing grasping in point clouds, gripper models are crucial to evaluate grasping stability.
Most works are now designed based on a specific type of grippers, and can hardly generalize to other grippers.
\cite{shao2020unigrasp} and \cite{xu2021adagrasp} proposed to encode gripper-specific features in the inputs and train gripper-specific grasp detectors.
They proved that by doing so, the model could learn to adapt to different types of grippers.

% TODO: xu2021adagrasp is based on depth images, not point clouds.
% here we may consider to re-design the taxonomy, e.g., by different difficulties existing in end-to-end grasp synthesis.
%   1. slow speed (early e2e works).
%.  2. orientation needs to be further addressed for robustness.
%.  3. close-loop
%.  4. adaptation to different gripper types
%.  etc.....

\subsubsection{Pixel-level Grasp Map Synthesis}
\label{sec:pixel-grasp}

Different from grasp detection, grasp map synthesis is similar to image segmentation, where the output is usually represented by a set of heat maps, indicating where and how to grasp.
One thing needed to be clarified is that there is no clear gap between grasp detection and grasp map synthesis.
One can imagine that in grasp detection, the dense estimation of grasp quality (e.g. in \cite{chu2018real, zhang2019real, zhao2021regnet}) for each pixel on image features is also a kind of grasp map synthesis, which is even more representative, but with a smaller-size output compared to the input.
Such methods could be seen as a transition between grasp detection and grasp map synthesis.

For pixel-level grasp map synthesis, transfer from segmentation algorithms is also widely explored.
For example, U-Net \cite{ronneberger2015u} has been widely used for grasp map synthesis \cite{liu2019active, shao2019combining, shao2019suction, chalvatzaki2020orientation}.
Such an encoder-decoder architecture is widely used to synthesize pixel-wise grasps \cite{asif2019densely, kasaei2021mvgrasp, cao2021suctionnet, wang2021graspness, xu2021gknet}.
Another similar formulation for pixel-wise grasp synthesis is called grasp manifolds, proposed by \cite{hager2021graspme}.
Since grasp map is more informative and could provide a global grasp affordance which indicates the grasp quality of the current viewpoints, it enables selection of the best view \cite{kasaei2021mvgrasp, wang2021graspness}, provided the assumption that the camera is not fixed, which holds in most cases for robots.
It is defined by a close set of points on objects representing graspable areas.
Besides, with the mobility of robots, the interaction could be imposed on the workspace to actively clear out around the graspable area \cite{zeng2018learning, deng2019deep, liu2019active} when no grasps are available.
Also, as mentioned above, reactivation is needed to recover from failures, and some works have explored reactive grasping policy learning based on pixel-level grasp maps \cite{morrison2018closing, morrison2019multi, morrison2020learning}.

\section{Object-centric Grasp Synthesis}
\label{sec:semantic-grasp}

In real applications, grasping usually serves for more complicated tasks requiring object-centric perception. 
Developments of learning methods make it possible to integrate the understanding of high-level concepts while executing grasping. 
In this section, we are going to review recent algorithms based on object-centric semantics.

\subsection{Overview}

Different object-centric semantics should be considered under different situations.
In this paper, we will discuss three types of them:
\begin{itemize}
%	\item {\bf Task-oriented Grasp Synthesis:} Grasping always relates to object functions and tasks. Task-oriented, also called task-driven, task-specific or functional, grasp synthesis was proposed under this background. It is still a challenging yet practical problem for intelligent and autonomous robots.
	\item {\bf Object-specific Grasp Synthesis:} Object-specific grasp synthesis aims to retrieve and grasp objects belonging to a specific class in clutter scenes. To specify a target, a class name is usually specified as the condition of grasping.
	\item {\bf Interactive Grasp Synthesis:} Interactive grasp synthesis means specifying targets using natural languages, which includes richer information about objects including attributes and relationships with other objects. It is worth noting that interactive grasping is different from interactive perception in robotics, which in most cases means perception based on interaction with environments \cite{bohg2017interactive}. We also included some works related to grasp synthesis based on interactive perception in \secref{sec:pixel-grasp}.
	\item {\bf Relational Grasp Synthesis:} Relational grasp synthesis is needed when grasping may have a possible negative effect on other objects. Planning algorithms \cite{lavalle2006planning} are feasible to handle such situations given environment models. With deep learning, the model-free understanding of object relations has also been explored in recent years. 
\end{itemize}
All types of object-centric grasping methods are built on top of robust grasp synthesis algorithms, and the difference lies in the introduction of semantics, which, to some extent, is parallel to grasp synthesis.
The motivation behind this is that we want robots to understand the world as a human can do, interact with humans in a natural way, and finish complicated tasks autonomously and robustly, which has been pursued for decades by almost all roboticists.

\subsection{Object-Specific Grasp Synthesis}

Object-specific grasp synthesis involves visual semantics of objects. 
When executing grasping, robots need to associate synthesized grasps to object instances, clearly be aware of which object it is going to grasp and how to grasp, and if possible, avoid possible collisions with other objects.

Methods based on template matching introduced in \secref{sec:template-match} are possible candidates to build associations between grasps and objects given grasp demonstrations or a grasp planner, though partial observability of object models may have negative effects on final performance.
To handle partial observability, reconstruction methods could be used to recover unseen parts of objects \cite{glover2008probabilistic, yang2021robotic, agnew2021amodal}, or a better view could be explored for better grasps \cite{chen2020transferable}.

Another way is following a recognize-first-then-grasp workflow.
By breaking down the problem into two independent components, advanced methods could be integrated as solutions \cite{zhang2018robotic}.
However, object-specific grasping is usually needed in dense clutter, in which objects may occlude and overlap each other severely, invalidating such naive matching methods.
To solve this problem, \cite{zeng2018robotic} proposed grasp-first-then-recognize, which avoids associations between detected objects and grasps with an acceptable loss of efficiency.
Nevertheless, when a specified object is requested with many other disturbance terms, such methods will be costly.

It is also possible to directly bind grasps onto objects when synthesizing grasping.
It can be achieved by combining grasp detection with object recognition \cite{guo2016object, jang2017end, zhang2019roi} or semantic segmentation \cite{asif2017rgb, chen2020towards, murali20206, dong2021mask, dong2021real, li2021simultaneous, ainetter2021end, ainetter2021depth}.
An alternative way is using reinforcement learning to encourage actions of grasping a specified object \cite{iqbal2020toward}.
Advantages of such methods include faster inference speed and higher accuracy.

One main drawback of the above methods is that they usually sacrifice generality to unknown objects since they cannot be recognized by most object detectors or scene segmentation algorithms.
One possible way to solve this problem is leveraging recent unseen object instance segmentation methods \cite{xiang2020learning, xie2021unseen}, which makes it possible to remove unknown objects in clutter for getting targets \cite{sundermeyer2021contact}.
Nonetheless, it still cannot recognize the semantics of unknown objects.
Another way is to involve human interventions for online learning \cite{kasaei2021simultaneous}.
Such methods require expert knowledge, which is expensive to get.

\subsection{Interactive Grasp Synthesis}

Similar to object-specific but more challenging, interactive grasp synthesis also requires an understanding of visual semantics by interaction with humans.
Benefiting from advances in visual-linguistic grounding \cite{nagaraja2016modeling, hu2017modeling, yu2018mattnet, lu2019vilbert, su2019vl, chen2020uniter}, natural languages could be the interface of interaction based on visual observations.
% For convenience and consistency with previous works, we call human commands ``referring expression'' (RP).
One advantage of interactive grasping is that natural languages always include much richer semantics than a simple word, and therefore, it is possible for robots to recognize and grasp unknown objects by their attributes or spatial relationships with other objects \cite{guadarrama2014open}.
To keep this survey consistent, we only review works directly relating to robotic grasp synthesis.

Most works focused on building interactive grasping systems with separated components.
In terms of grounding spatial relationships, \cite{guadarrama2013grounding} grounded spatial relationships by modeling it as a multi-class logistic regression problem by taking a preposition and referential object as input.
\cite{alomari2017natural} proposed a method based on Robot Control Language \cite{matuszek2013learning} to ground natural languages for robotic manipulation.
Alternatively, \cite{paul2018efficient} proposed a probabilistic model to handle abstract concepts, like ``one'', ``two'' or ``the first'', ``the second''.

With the help of deep learning, grounding performance and scalability have been further improved.
\cite{hatori2018interactively} used a simple multi-branch network for end-to-end grounding of target objects and destinations.
In their method, they firstly detected objects in their workspace by SSD \cite{liu2016ssd}.
After that, based on detected regions, they extracted visual features of objects and linguistic features of the given command by CNN and LSTM \cite{hochreiter1997long} respectively.
Finally, the network directly grounded referred objects and corresponding target positions.
\cite{shridhar2018interactive} presented a generative grounding method, named ``INGRESS'', in which objects are grounded by the similarity between the given command and a set of generated self-referential \cite{johnson2016densecap} and relational captions \cite{nagaraja2016modeling}.
It also allows robots to ask questions when ambiguity is detected.
Later, they expanded this work with a POMDP planner for decision-making between interaction and grasping \cite{shridhar2020ingress}.
\cite{cohen2019grounding} modeled a shared space for visual attributes and linguistic concepts, and grounded objects by similarities.
Besides visual-linguistic inputs, \cite{wang2021audio} introduced additional audio information of objects to finish grounding, since sometimes only visual information is not sufficient (e.g. an opaque bottle with different substances in it).
Followed by a grasp planner, such methods could grasp objects specified by attributes or relationships with other objects.
However, the overall pipeline of these systems is similar to the recognized-first-then-grasp workflow, and hence, it can hardly transfer to dense clutters.
Actually, most of them assumed objects are scattered.
Therefore, to handle clutter scenes, discriminative models were demonstrated to be more effective by \cite{zhang2021invigorate}.
In their paper, they proposed INVIGORATE, an interactive visual grounding and grasping method based on POMDP, with visual and linguistic observations and grasp-sequence actions.

End-to-end methods are also explored recently based on visual and linguistic inputs.
\cite{chen2021joint} presented a simple multi-modality network with ResNet-based \cite{he2016deep} visual branch and LSTM-based \cite{hochreiter1997long} linguistic branch, which directly regresses a suitable grasp for specified targets as output.
To train their models, they re-labeled Visual Manipulation Relationship Dataset \cite{zhang2018visual} with natural language commands, which is labor-exhaustive.
To prevent labeling large-scale datasets, \cite{lynch2021language} proposed to train policies in a semi-supervised way with a small amount of labeled data ($<1\%$).
They also took advantage of large-scale pre-trained language models \cite{devlin2018bert, radford2019language}.
They demonstrated that the final goal-conditioned policy performed extremely well.
\cite{shridhar2021cliport} proposed CLIPort, also based on a large-scale unsupervised representation learning model named CLIP \cite{radford2021learning}, which demonstrated surprisingly good generality when trained with a few demonstrations for natural-language-conditioned robotic manipulation.

\subsection{Relational Grasp Synthesis}

Relational grasp synthesis means that when grasps are being executed, relations among objects should be considered to plan for a grasping sequence in clutter.
This planning is necessary because improper order of grasping will result in irrevocable damages to objects.
In this section, we will review feasible solutions for this problem.
%Methods included here are categorized into two classes: 1) grasp planning based on object relationship understanding; 2) task planning given environment dynamics.

%\subsubsection{Task Planning in Clutter}
%
%Task planning \cite{bercher2019survey} is a general framework for solving long-term complex robotic problems.
%It usually builds a symbolic planning space based on known environment models.
%
%\subsubsection{Relationship Recognition}

Planning algorithms such as task planning are possible candidates for solving relational grasping problems \cite{garrett2021integrated}.
However, they usually require environment dynamics to plan.
For grasping in clutters with possible piles of objects, support relationships can represent simplified dynamics of objects.
\cite{panda2013learning} presented a learning-based segment-first-then-recognize approach for support relation analysis.
Interestingly, they demonstrated that the rule-based approach is better than the proposed learning-based approach.
Actually, assuming that objects are convex and their models and poses are known, support relationships can indeed be synthesized by geometric and physical analysis \cite{mojtahedzadeh2015support, kartmann2018extraction}.
In \cite{mojtahedzadeh2015support}, they proposed 4 basic types of support relations and solved them using static equilibrium analysis (SEA) and learning methods when objects are fully and partially detected respectively.
In \cite{kartmann2018extraction}, they adapted the (SEA) approach from \cite{mojtahedzadeh2015support} to single-view point with incomplete object models.
Later, they tried to select the best view for SEA \cite{grotz2019active} or consider object uncertainty of types and shapes \cite{paus2021probabilistic}, which improved performance in real scenarios.
However, as mentioned, such methods rely on assumptions that either object models are convex and fully accessible or detected objects could be approximated by convex shapes, limiting their practicality in real scenarios.

To get rid of these assumptions, visual relationship understanding \cite{lu2016visual} in vision also provides insights to grasp planning in clutter.
\cite{lu2016visual} demonstrated that object pair features from deep learning could be directly used for high-performance semantic relationship understanding of arbitrary objects.
Inspired by visual relationship detection, \cite{zhang2018visual} presented the concept of {\it visual manipulation relationship} (VMR), which is similar to the support relationship but defined on purely visual features.
They designed an end-to-end network for the detection of VMRs and demonstrated that deep learning could be directly used for the classification of VMRs among non-convex objects.
Further, they extended this work with more object detectors and showed that advanced approaches in object detection and visual relationship detection could be directly transferred for better performance of VMR analysis \cite{zhang2020visual}.
Recently, graph neural networks (GNNs) are widely proved to be efficient for detecting relationships among objects \cite{wu2020comprehensive, zhang2020deep}.
\cite{zuo2021graph} proposed a GNN-based method for VMR detection and achieved better performance.
Such kind of methods has been successfully applied to the decision of grasping sequence in dense-clutter scenes \cite{zhang2019multi, park2020single, zhang2021invigorate}.

\section{Open Problems}
\label{sec:future}

From the discussions above, it is obvious that the development of grasping is from structural to intelligent.
Early works mostly focused on mechanical analysis and optimality, which requires full models of objects, grasps, and environments.
Later, learning is applicable to relax assumptions on environments, which enables the deployment of grasping algorithms in partially unstructured scenarios.
Recently, most works are exploring grasping with high-level visual concepts, aiming for adaptation to daily home scenarios, though currently, it is still far behind this final goal.

There are also some interesting points which are worth noting.

Firstly, there is no doubt that mechanics should be directly responsible for the stability of grasping.
However, most current works focus on heuristic methods based on learning and show surprisingly good performance on grasp synthesis on unknown objects.
Though investigated by some works already (e.g. \cite{mahler2017dex, liang2019pointnetgpd, ni2020pointnet++}), there is still a gap between these two domains, i.e., analytic grasp synthesis and data-driven grasp synthesis.
So the problem is {\it could we take the best of both analytic and data-driven to develop robust and scalable grasping methods?}
One way we believe promising is to learn intuitive physics \cite{kubricht2017intuitive, ye2018interpretable, riochet2018intphys} for stable grasping.
It comes from the observation of how our humans grasp objects.
In most cases, we implicitly roughly infer some physical properties of objects before grasping so as to avoid failures, e.g., the friction coefficient, center of mass, and 3D geometry of unseen parts based on our knowledge base.
With these rough models, we then implicitly plan a reasonable (which may not be optimal) grasp.
Such a pipeline is similar to sample-based methods introduced in \secref{sec:sample-grasp}, most of which only focus on local geometric information instead of object-level physical properties.
Another alternative way to involve consideration of physics is using physical simulators \cite{battaglia2013simulation}, especially with the help of recent advances in differentiable simulators \cite{hu2019chainqueen, hu2019difftaichi, werling2021fast}.

Secondly, scene understanding with high-level semantics is also closely related to robotic manipulation tasks, which has also been actively explored in grasp synthesis and is critical for robot intelligence.
Currently, the main difficulties are generalization to open-set objects and worlds.
Recent progress in unsupervised representation learning shows that it is promising to learn structured representations for unknown objects and concepts \cite{erhan2010does, radford2021learning}.
% Besides, such methods always support multi-modal inputs, which are also crucial for stability analysis of grasps.
Therefore, the problem is {\it could we take advantage of large-scale unsupervised representation learning for developing robust grasping skills in semantic scenarios?}
To grasp open-set targets, one straightforward way is to train grasp policies directly based on the large-scale pre-trained models.
Also, one may consider a composable alternative, in which semantics could be analyzed first, and grasps could be then synthesized in an object-centric way.
To do so, a robust object-centric grasp detector is needed.
Another important thing is relationship understanding with open-set objects.
Humans can retrieve and grasp targets efficiently in daily scenes even with unrecognized distractors.
This ability is built on top of the hierarchical understanding of semantics.
For example, a command ``fetch me the red bottle on the dinner table'' will involve a two-layer relationship ``dinner room - dinner table - red bottle'', where the first relationship ``dinner room - dinner table'' is implicit and based on prior semantic knowledge.
Such tasks are still challenging for robots to complete.

Finally, uncertainty is everywhere in practice.
In traditional robotics, it is critical to handle uncertainty for planning and control.
However, most learning-based grasping approaches simply utilize one-shot greedy inference models.
Thus, it is meaningful to consider {\it could we model the uncertainty from the learned models when planning grasps for robustness?}
To some extent, neural networks are like sensors, providing high-level noisy semantic information for decision making.
We believe that one-shot greedy policies are not the optimal way to use these observations.
To consider model uncertainty, the first thing needed to be handled is how to model reasonable uncertainty for outputs of neural networks.
Model calibration \cite{guo2017calibration} is a useful tool to calibrate output uncertainties.
With uncertainties, better decisions could be made to optimize final goals with given constraints (safety, success rate, etc.).
Since decision-making in robotics usually involves a sequential decision-making problem with partial observations, historic information is also helpful to optimize actions.
Partially observable Markov Decision Process (POMDP) \cite{monahan1982state} is a natural candidate to consider uncertainty, long-term decision making, and partial observability in a principled way.
Recent advances also illustrated promising results for POMDP to solve large-scale problems \cite{ye2017despot, garg2019learning}.

\section{Conclusions}
\label{sec:conclusion}

In this paper, we review the history of robotic grasp synthesis approaches, including analytic methods, data-driven methods, and recent object-centric methods.
Analytic methods are usually based on top of known object models and mechanical analysis, which can theoretically ensure stability but with strong assumptions and simplifications limiting its application in practical scenarios.
Data-driven methods are derived from neuropsychology and are mostly heuristic. However, it relaxed the assumptions made by analytic methods and hence, is widely used in real-world scenarios.
In particular, benefitting from recent progress in learning techniques, it achieves commendable performance in grasping tasks and is promising to play important roles in robotic autonomy.
Recently, with developments in semantic vision, object-centric methods have been more and more actively investigated.
Object-centric methods combine the understanding of object semantics with grasp synthesis for semantic grasping, which is more close to our daily life instead of industrial applications.
We believe that in the future, vision-based intuitive physics, open-set grasping with semantic representations, and planning under partial observability and uncertainty will be the future trends for robotic grasping.

\section*{Author Contributions}

Hanbo Zhang finished most parts of this manuscripts. Jian Tang and Shiguang Sun helped to collect and organize the literature. Xuguang Lan is the supervisor of Hanbo Zhang, Jian Tang, and Shiguang Sun, and he is also the corresponding author and responsible for all contents.

\bibliographystyle{plain}
\bibliography{fair}

\end{document}